\documentclass{article}

\PassOptionsToPackage{numbers, compress}{natbib}
\usepackage[preprint]{neurips_2025_ml4ps}

\usepackage[utf8]{inputenc} 
\usepackage[T1]{fontenc}    
\usepackage{hyperref}       
\usepackage{url}            
\usepackage{booktabs}       
\usepackage{amsfonts}       
\usepackage{nicefrac}       
\usepackage{microtype}      
\usepackage{xcolor}         
\usepackage{amsmath}
\usepackage{amssymb}
\usepackage{mathtools}
\usepackage{amsthm}
\usepackage{graphicx}

\usepackage{enumitem}
\usepackage{siunitx}
\usepackage{amsfonts}
\usepackage{nicefrac}
\usepackage{lipsum}
\usepackage{multirow}
\usepackage{color}
\usepackage{xspace}
\usepackage{soul}
\usepackage{listing}
\usepackage{amsmath}
\usepackage{mathtools}
\usepackage{multirow}
\usepackage{listings}
\usepackage{color}
\usepackage{here}
\usepackage{import}
\usepackage{bm}
\usepackage{adjustbox}
\usepackage{subcaption}
\usepackage{placeins}
\usepackage{siunitx}
\usepackage{tabularx}
\usepackage{multirow}
\usepackage{makecell}

\usepackage[capitalize,noabbrev]{cleveref}

\title{
JetFormer: A Scalable and Efficient Transformer for Jet Tagging from Offline Analysis to FPGA Triggers
}

\author{
  Ruoqing Zheng$^1$ \quad
  Chang Sun$^2$ \quad
  Qibin Liu$^3$ \quad
  Lauri Laatu$^1$ \quad
  Arianna Cox$^1$ \quad \\
  \textbf{
  Benedikt Maier$^1$ \quad 
  Alexander Tapper$^1$ \quad
  Jose G. F. Coutinho$^1$ \quad
  Wayne Luk$^1$ \quad
  Zhiqiang Que$^{1}$\thanks{Corresponding author and project advisor: z.que@imperial.ac.uk.}~~}
  \\ [1.0em]
  $^1$Imperial College London, UK \\
  $^2$California Institute of Technology, USA \\
  $^3$SLAC National Accelerator Laboratory, USA \\
}

\begin{document}
\maketitle

\begin{abstract}
We present JetFormer, a versatile and scalable encoder-only Transformer architecture for particle jet tagging at the Large Hadron Collider (LHC). Unlike prior approaches that are often tailored to specific deployment regimes, JetFormer is designed to operate effectively across the full spectrum of jet tagging scenarios, from high-accuracy offline analysis to ultra-low-latency online triggering. The model processes variable-length sets of particle features without relying on input of explicit pairwise interactions, yet achieves competitive or superior performance compared to state-of-the-art methods. On the large-scale \textsc{JetClass} dataset, a large-scale JetFormer matches the accuracy of the interaction-rich ParT model (within 0.7\%) while using 37.4\% fewer FLOPs, demonstrating its computational efficiency and strong generalization. On benchmark HLS4ML 150P datasets, JetFormer consistently outperforms existing models such as MLPs, Deep Sets, and Interaction Networks by 3–4\% in accuracy. To bridge the gap to hardware deployment, we further introduce a hardware-aware optimization pipeline based on multi-objective hyperparameter search, yielding compact variants like JetFormer-tiny suitable for FPGA-based trigger systems with sub-microsecond latency requirements. Through structured pruning and quantization, we show that JetFormer can be aggressively compressed with minimal accuracy loss. By unifying high-performance modeling and deployability within a single architectural framework, JetFormer provides a practical pathway for deploying Transformer-based jet taggers in both offline and online environments at the LHC. Code is available at 
https://github.com/walkieq/JetFormer.
\end{abstract}

\section{Introduction}
\label{sec:introduction}

Machine learning has become a powerful tool in the analysis of high-energy physics (HEP) data. At the CERN Large Hadron Collider (LHC)~\cite{Lyndon_Evans_2008}, protons are accelerated and collided at high energy, producing jets, which are collimated sprays of particles. Jet tagging is the process of classifying jets based on the type of particle that initiated them. This task is essential for isolating signatures of heavy particles such as Higgs bosons or top quarks from the overwhelming background of light-flavor quark and gluon jets~\cite{qu2022particle}.

In recent years, deep learning techniques have led to significant advances in jet tagging performance. Among these, transformer-based architectures have reached state-of-the-art performance, owing to their ability to capture global relationships among jet constituents via self-attention. Transformers often outperform CNN-, GNN-, and RNN-based approaches in accuracy and generalization. A notable example is the Particle Transformer ParT~\cite{qu2022particle}, which incorporates particle interactions through a customized attention mechanism and achieves leading performance across multiple jet-tagging benchmarks.

During high-energy collisions at the LHC, particle detectors such as ATLAS~\cite{The_ATLAS_Collaboration_2008} and CMS~\cite{The_CMS_Collaboration_2008} record billions of proton-proton collisions every second. Among these events, only a very small fraction correspond to signals of interest. Efficiently identifying the interesting signals in a noisy environment is a major experimental challenge. Since it is neither technically nor economically feasible to record all collision data, the data rate must be reduced in real-time. This reduction is achieved through a multi-level trigger system, whose first stage is implemented in custom hardware such as the Level-1 trigger system (L1T) at CMS. The L1T rapidly selects potentially interesting events for further downstream analysis while discarding the vast majority of events, reducing the event rate to a level that can be handled by the subsequent software-based trigger and offline storage. Given the limited capacity of the detector buffers, the L1T decision-making process is subject to stringent timing constraints. Hence, the L1T operates on Field Programmable Gate Arrays (FPGAs), enabling low latency as well as high data throughput and custom design~\cite{The_CMS_Collaboration_2008}. As a result, the L1T design requires careful optimization to achieve maximal physics performance within tight constraints on latency and hardware resources.

In the CMS Phase-2 upgrade, the L1T will see significant increases in FPGA resources and a more generous latency budget, which allows for more sophisticated physics reconstruction algorithms. In addition, the inclusion of track information will allow the L1T Correlator trigger to create particle-level features and enable jet reconstruction. This opens the door to deploying jet tagging algorithms directly in the L1T to improve event selection efficiency under extreme pileup conditions. Given the strong performance of transformer-based architectures for offline jet tagging, there is strong motivation to adapt and optimize transformer models for implementation on FPGAs. However, large model size and structural complexity pose challenges for compression and deployment on resource-constrained platforms such as FPGAs. Compact transformer models~\cite{wojcicki2022accelerating} have been successfully compressed and deployed on FPGAs, but often require manual and custom quantization strategies and hardware-aware modifications. Translating neural network models into high-level synthesis (HLS) code suitable for FPGA deployment is typically performed using tools such as \texttt{hls4ml}, an open-source library that supports a variety of network types~\cite{aarrestad2021fast, ghielmetti2022real}. However, \texttt{hls4ml} currently does not support all transformer models. In this work we explored an alternative framework, Allo, which offers more flexibility for deploying transformer-based models on FPGAs.

In this work, we introduce JetFormer, a versatile encoder-only Transformer architecture designed for jet tagging across the full range of high-energy physics (HEP) use cases, from high-accuracy offline analysis to ultra-low-latency online triggering. JetFormer processes unordered sets of particle features without relying on interaction terms, yet achieves state-of-the-art or competitive performance on both small-scale HLS4ML 150P benchmarks and the large-scale \textsc{JetClass} dataset. Notably, a full-capacity JetFormer matches the accuracy of the interaction-rich ParT model within 0.7\% while using 37.4\% fewer FLOPs, demonstrating its efficiency and scalability.

To enable deployment in resource-constrained environments such as FPGA-based trigger systems, where sub-microsecond latency is required, we develop a hardware-aware optimization and compression pipeline. This includes multi-objective hyperparameter tuning via Optuna, structured pruning, and aggressive 1-bit quantization. Pruning reduces computational cost by ~50\% with minimal (<0.5\%) accuracy loss, while quantization shrinks model size by 82–92\% at the cost of only 1.5–3.5\% absolute accuracy drop. Using custom extensions to the Allo high-level synthesis framework, we successfully synthesize both pruned and quantized variants (e.g., JetFormer-tiny) onto FPGA hardware, validating the feasibility of real-time Transformer inference in LHC trigger systems.

Critically, JetFormer is not limited to tiny configurations: its modular design supports seamless scaling, allowing the same architectural foundation to serve diverse latency and accuracy requirements. This work thus establishes a unified end-to-end framework, from scalable model design and systematic compression to hardware implementation, that bridges the gap between cutting-edge deep learning and practical deployment in HEP.

This work makes the following contributions:
\begin{itemize}
    \item Section~\ref{sec:jetformer}: 
    We propose JetFormer, a scalable, encoder-only Transformer for jet tagging that excels across model sizes, from compact variants for online triggers to larger instances for offline analysis, and introduce an automated pipeline for hardware-aware optimization and compression.
    \item Section~\ref{sec:allo}: We extend the Allo framework to support essential Transformer operations, enabling end-to-end FPGA synthesis of compressed JetFormer models. 
    \item Section~\ref{sec:results}: We present comprehensive evaluations of JetFormer’s accuracy, efficiency, and compressibility across multiple datasets. 
    
\end{itemize}

This work first introduces background knowledge and related work in Section~\ref{sec:bg}. It then presents the detailed implementation in Section~\ref{sec:jetformer}, \ref{sec:allo} and \ref{sec:results}, respectively. Finally, Section~\ref{sec:conclusion} concludes the current achievements and outlines directions for future improvement.

\section{Background and Related Work}
\label{sec:bg}
\subsection{Jet Tagging}
Machine learning methods have been extensively developed for jet tagging. Based on a variety of input representations such as images, graphs and sequences, models can be designed as convolutional neural networks (CNNs)~\cite{li2020attention}, graph neural networks (GNNs) ~\cite{gong2022efficient} or recurrent neural networks (RNNs)~\cite{khoda2023ultra}. Among GNN-based methods, JEDI-net~\cite{moreno2020jedi} is a notable jet tagging algorithm based on interaction networks. It models jets as graphs of constituent particles with pairwise interactions. Unlike traditional approaches, JEDI-net directly processes particle features without requiring any pre-processing or geometry assumptions. It achieves strong classification performance across multiple jet types, and therefore becomes widely used as a baseline for the jet tagging task. Alternatively, ParticleNet~\cite{qu2020jet} models jets as particle clouds, which are unordered sets of constituent particles. Built upon the dynamic graph convolutional neural network (DGCNN) framework, ParticleNet captures both local and global structures within jets. 

More recently, transformer-based architectures are proposed. The Particle Transformer (ParT)~\cite{qu2022particle} is a state-of-the-art (SOTA) model for jet tagging with a custom attention mechanism known as particle multi-head attention (P-MHA). This mechanism incorporates pairwise features as an attention bias. A related architecture has also been proposed that relies solely on pairwise features for computing attention, thereby lowering computational demands ~\cite{Wu_2025}. Additionally, SAL-T~\cite{wang2025spatially} presents a spatially-aware attention model that captures geometric relationships and spatial information while achieving reduced computational cost. These approaches further improves model performance on various datasets, demonstrating the potential of transformer-based designs in jet tagging task. However, these transformer-based models are typically too large for FPGA deployment in L1T. Instead, they are more suitable for offline uses, or potentially the high-level trigger systems after L1T, where latency budgets are less stringent. 

\subsection{Low Latency Jet Tagging on FPGAs}

Previous research has implemented Multi-Layer Perceptron (MLP) networks on FPGAs for jet tagging applications \cite{duarte2018fast, coelho2021automatic}, though these approaches achieved limited accuracy. Recent work \cite{que2022reconf, que2022opt, que2024ll} has shifted toward hardware-aware co-design of Graph Neural Networks (GNNs), incorporating optimization techniques like quantization, pruning, and efficient graph representations to satisfy the stringent latency and resource constraints of Level-1 Trigger (L1T) systems while maintaining model performance. Ref.~\cite{que2022reconf} presents the first FPGA-based implementation of the GNN-based JEDI-net~\cite{moreno2020jedi} for jet tagging. LL-GNN~\cite{que2024ll} adopts task-level parallelism, sublayer fusion, and latency-aware algorithm-hardware co-design, resulting in a sub-microsecond latency and initiation intervals.  
In addition, \cite{odagiu2024ultrafast} applies quantization-aware training and utilizes a uniform 8-bit fixed-point representation, evaluating deep sets (DS), interaction network (IN) and multilayer perceptron (MLP). 
More recently, MLP-Mixers (MLPM)~\cite{sun2025fast} have been introduced for jet tagging applications, utilizing the quantization-aware training approach HGQ~\cite{sun2026hgq}. 
In addition, JEDI-linear~\cite{jedi-linear}, a linearized variant of JEDI-net, has been proposed. This architecture also combines HGQ~\cite{sun2026hgq} quantization-aware training with distributed arithmetic (DA) optimization through da4ml~\cite{da4ml}, achieving superior model accuracy while significantly reducing hardware resource requirements. JEDI-linear currently represents the state-of-the-art (SOTA) solution for real-time jet tagging implementations on FPGAs.

Previous work for real-time jet tagging based on transformers has been done for high-level features in \cite{transformer_icl} and particle level inputs achieving $\mathcal{O}(1\mathrm{\mu s})$ in~\cite{transformer_uw, jiang2025low}. More recently in~\cite{submicrotransformer} two transformer architectures: Multi-Head Attention and Linformer, an architecture that reduces the complexity of computing the attention, that use particle-level inputs achieving a latency of $\mathcal{O}(100\mathrm{ns})$.

\subsection{Allo}
Allo is a hardware customization framework built on the Multi-Level Intermediate Representation (MLIR) compiler infrastructure~\cite{mlir}, which represents models using extensible \textit{dialects} and allows transformations from PyTorch/Hugging Face programs down to LLVM IR or HLS C/C++ for CPUs and FPGAs~\cite{chen2024allo}. It introduces custom MLIR dialects to describe computation, memory access, communication, and data types, and decouples the algorithm from hardware-specific optimizations such as pipelining. In the frontend, neural network operators (e.g., linear, softmax, convolution) are defined in a declarative DSL with parameterized types and shapes, serving as reusable kernel templates that are lowered to MLIR. The backend then schedules and optimizes these IR kernels and generates code for CPU simulation or Vitis HLS, supporting software emulation, hardware emulation, and full hardware execution (though in this work we limit validation to CPU, \texttt{sw\_emu}, and \texttt{hw\_emu} due to resource and time constraints).

\section{JetFormer}
\label{sec:jetformer}

This section introduces JetFormer, a compact transformer architecture specifically designed for jet tagging in high-energy physics. The focus is on both architectural adaptations that make JetFormer suitable for hardware deployment, and the hardware-aware pipeline that optimizes and compresses the model. We first present the JetFormer architecture and its modifications on top of common transformers. Next, the datasets used for training and evaluation are introduced. The section then explains the implementation pipeline in detail, including dataset preparation, training setup, hyperparameter optimization, and compression methods such as pruning and quantization.

\subsection{JetFormer Architecture}
\begin{figure}[ht]
    \centering
    \begin{subfigure}{0.5\textwidth}
        \centering
        \includegraphics[scale=0.8,trim=0 1.2cm 0 1cm, clip]{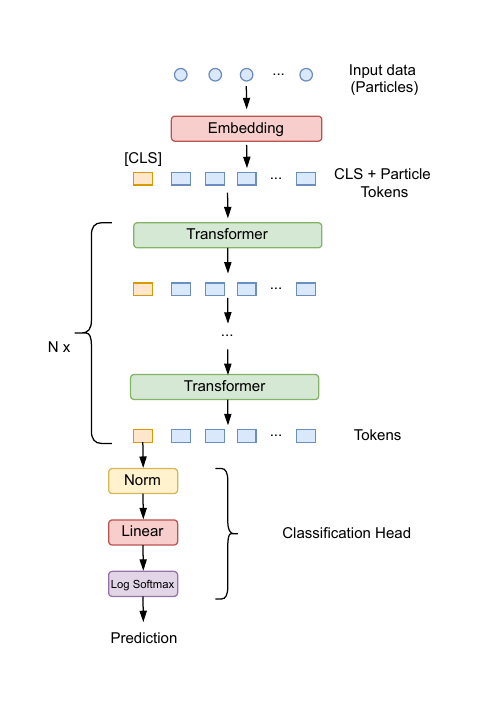}
        \caption{Overall model architecture.}
        \label{fig:model_arch}
    \end{subfigure}%
    \hfill
    \begin{subfigure}{0.5\textwidth}
        \centering
        \includegraphics[scale=0.9]{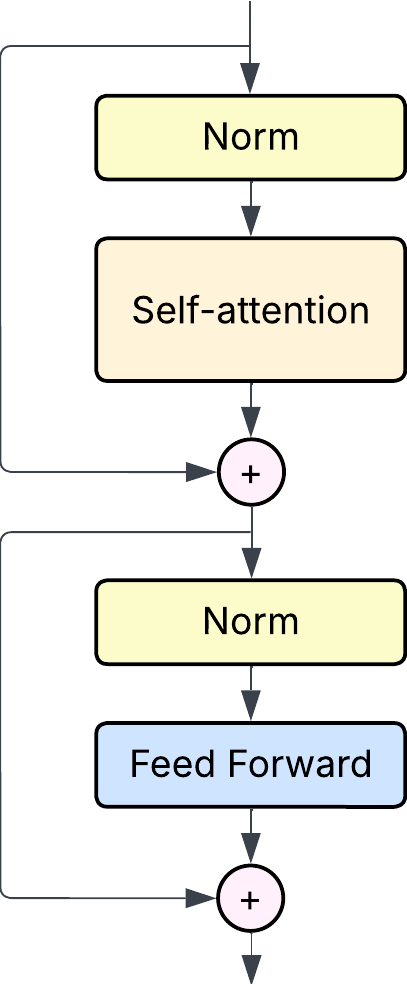}
        \caption{Transformer block architecture.}
        \label{fig:attention_arch}
    \end{subfigure}
    \caption{The model architecture and transformer block design.}
    \label{fig:combined}
\end{figure}
The JetFormer is based on an encoder-only Transformer architecture~\cite{vaswani2017attention}, inspired by Bidirectional Encoder Representations from Transformers (BERT)~\cite{devlin2019bert}, a foundational encoder-only model widely used in natural language processing (NLP). For classification tasks such as sentiment or sentence categorization, BERT prepends a special \texttt{[CLS]} token to the input sequence, whose final hidden state serves as a contextualized summary of the entire sequence and is fed into a lightweight classification head. 
The use of a dedicated token to produce a sequence-level representation for classification directly motivates our approach to jet tagging, as this task similarly requires assigning a single physical label to an entire jet, which is composed of an unordered set of particles.

As shown in Figure~\ref{fig:model_arch}, JetFormer consists of three main modules: input embedding, several transformer blocks, and a classification head as the output layer. Following a similar input representation scheme in BERT, each particle in the jet is treated as an input token, while each jet (sequence of particles) is modeled as a sequence. Each token in the sequence is projected to a vector in the embedding layer. The positional encoding is excluded to preserve the permutation invariance of the JetFormer. 
A learnable class token, analogous to BERT’s \texttt{[CLS]}, is attached to the particle sequence. This token is randomly initialized and optimized end-to-end to capture global, jet-level features through interactions with all particles.

The resulting embedding sequence is then processed by multiple stacked Transformer blocks, each comprising multi-head self-attention and feedforward sublayers with residual connections and layer normalization. These blocks enable the model to implicitly learn complex particle-particle relationships and contextual representations without explicit feature engineering. After the final Transformer block, the hidden state corresponding to the class token is extracted and passed to a classification head. The output layer applies a log-softmax function to produce the final probability distribution over jet classes. Throughout the network, layer normalization is employed to stabilize training, improve generalization, and ensure consistent feature scaling across layers.

Figure~\ref{fig:attention_arch} illustrates the structure of a single transformer block. The key module in the block is the multi-head self-attention. A two-layer feedforward network (FFN) is added after the self-attention layer, with a layer normalization before each layer. SiLU~\cite{elfwing2017slu} is adopted as the activation function which is a combination of both Sigmoid and ReLU. 
In each block, two residual connections are also added to connect the input of the block to the self-attention output, and to connect the input of the FFN to its output.

For easier hardware implementation, two major changes are made. All layer normalization in the model is substituted with batch normalization. Compared to layer normalization, which needs to compute mean and variance at runtime, the parameters of batch normalization are precomputed and fixed during inference~\cite{pmlr-v37-ioffe15}. This makes the normalization process much easier to compute and deploy on hardware. Another change is the replacement of SiLU with ReLU, which avoids several computationally expensive operations like exponential and division.

\subsection{Training}
\label{sec:training}
For training, the AdamW~\cite{loshchilov2017decoupled} optimizer is employed with $\text{weight decay}=0.01$, and initial $\text{lr}=0.001$. Three learning rate schedulers are evaluated:
\begin{itemize}
    \item \textbf{ReduceLROnPlateau}: Reduces the learning rate when a metric stops improving for consecutive epochs. With learning rate $\text{decay factor}=0.5$, $\text{patience}=2$ and $\text{minimum lr}=10^{-4}$, the validation loss is selected as the monitored metric.
    \item \textbf{CosineAnnealingLR}: Gradually decreases the learning rate following a cosine function for each iteration. The max iteration \texttt{T\_max} is set to the number of training epochs. 
    \item \textbf{OneCycleLR}: First increases the learning rate to a predefined maximum, then decreases it combined with an annealing strategy. The proportion of the cycle spent increasing the learning rate is set to 0.2, with maximum learning rate of 0.001, and the annealing strategy is cosine.
\end{itemize}

Among these, OneCycleLR achieves the best validation performance and fastest convergence. Therefore, it is adopted as the final training configuration. A batch size of 256 is used for the 150-particle dataset, and 128 for the \textsc{JetClass} dataset. To avoid overfitting, early stopping is implemented to stop training and save the best model when validation performance no longer improves. 

\subsection{Hyperparameter Optimization (HPO)}
\label{sec:hpo}
This work performs hyperparameter optimization of transformer models trained on the 8-particle, 3-feature subset of the 150-particle dataset.
Optuna~\cite{akiba2019optuna}, an HPO framework is employed due to its lightweight design and efficient optimization algorithms. For hardware deployment, the model should achieve an acceptable performance with a compact size. Therefore, multi-objective optimization strategy is implemented to maximize model accuracy while minimizing FLOPs. The optimization process also evaluates and compares three different Optuna samplers. The best model identified in the HPO pipeline is denoted as JetFormer-tiny and is deployed in Section~\ref{sec:allo}.

\subsubsection*{Optuna HPO Workflow}
Three samplers (NSGAIISampler, TPESampler, BoTorchSampler) in Optuna are first compared and evaluated, and a suitable sampler is then selected based on performance for the final HPO process. The optimization is conducted using the 8-particle 3-feature dataset. The training-to-validation split ratio is 9:1. For each trial, a subprocess is initiated to evaluate a specified parameter configuration. The resulting validation accuracy and FLOPs are recorded in a SQLite~\cite{sqlite2020hipp} database for subsequent analysis and model selection.

\paragraph{Optimization settings.} The hyperparameter space is explored as follows:
\begin{itemize}
    \item \texttt{num\_transformers}: Number of transformer blocks, from integer 1 to 6.
    \item \texttt{dim\_heads}: Combination of the embedding dimension and number of attention heads in each transformer. To improve search efficiency and provide a feasible design space, the combinations are predefined: (8,2), (16,2), (32,2), (64,2), (64,4), (128,2), (128,4), (128,8). Each configuration ensures the dimensionality per head is at least 4 even for a small model, so that each attention head can learn meaningful feature patterns.
    \item \texttt{dropout}: Since the model does not show significant overfitting, only dropout rates of 0.0 and 0.05 are considered.
\end{itemize}

During each hyperparameter trial, the model is trained using the same optimizer and LR schedule as in the final model training: AdamW and OneCycleLR. Each trial is run with a batch size of 256 for up to 25 epochs. Most models converge within this range. To reduce unnecessary computation, early stopping with a patience of 4 epochs is applied.

\paragraph{Constraints.} An objective constraint is used to filter out less optimal models. Specifically, models with validation accuracy below 0.65 are considered “infeasible”, and therefore will be excluded from the Pareto front. Only models meeting a minimum performance threshold are retained during multi-objective optimization.

\paragraph{Sampler comparison.} Since both TPESampler and BoTorchSampler are based on Bayesian optimization, a number of startup trials (or warm-up steps) are required to construct the probability distributions or build a surrogate model. Therefore, during the initial phase, both samplers rely on a random sampling strategy. However, NSGAIISampler does not have any extra warm-up steps. Thus, for fair comparison, each sampler is evaluated for 80 trials:
\begin{itemize}
    \item \textbf{NSGAIISampler}: 80 trials with NSGAIISampler. 
    \item \textbf{TPESampler}: 20 startup trials with a random sampler, 60 trials with TPESampler.
    \item \textbf{BoTorchSampler}: 20 startup trials with a random sampler, 60 trials with BoTorchSampler.
\end{itemize}

\subsubsection*{Evaluation Metrics}
To compare the performance of the three samplers, the following evaluation metric is used:
\begin{itemize}
    \item \textbf{Hypervolume (HV) @ N trials}: Hypervolume~\cite{guerreiro2021hypervolume} measures the volume formed by the optimal solutions in the Pareto front. Better solutions or more diverse solutions can both lead to a larger HV. Hence, it quantifies both convergence and diversity of the Pareto solutions. A higher HV indicates better overall performance. Here, HV is evaluated at 40, 60, 80, and 100 trials. Since in our case, accuracy and FLOPs have opposite optimization directions (maximize accuracy and minimize FLOPs), accuracy is negated before computing HV, so that both objectives are consistently treated as maximization problems.
\end{itemize}

For later model compression and deployment, the model with the lowest FLOPs and an accuracy higher than 0.65 is selected and referred to as JetFormer-tiny.

\subsection{Pruning}
This work adopts a structured pruning pipeline based on the \texttt{torch-pruning} library~\cite{fang2023depgraph}, which builds a dependency graph so that when parameters are pruned in early layers, all dependent parameters in later layers are consistently removed, avoiding dimensional mismatches. We focus on structured pruning, which removes entire channels, filters, or groups, as opposed to unstructured pruning, which drops individual weights and typically achieves higher compression at the cost of irregular sparsity that is less hardware-friendly and often requires specialized support~\cite{10330640}. Weight importance is estimated using common criteria such as L1/L2 norms and Taylor expansion; in this work we adopt Taylor importance~\cite{molchanov2019importance}, which approximates the loss change after pruning using first-order expansion and combines weight magnitude and gradients. The pipeline prunes linear layers in JetFormer (input embedding, attention projections for keys/queries/values, and feedforward layers) in multiple steps: at each step, Taylor scores are computed on mini-batches, a fraction of parameters is removed according to a global pruning ratio (0.5, targeting roughly a 50\% FLOP reduction), and the model is fine-tuned (5 epochs per step, 5 steps total) for accuracy recovery. Effectiveness is assessed by comparing FLOPs, parameter count, validation accuracy, and inference time before and after pruning. Inference is performed on an NVIDIA RTX 3090 GPU with a large batch size (10,240) to reach 100\% utilization. Each measurement is repeated 500 times following 20 warm-up runs. The results include average latency and relative metric changes are reported.

\subsection{Quantization}
To further reduce the model size and computational cost, this work adopts 1-bit quantization~\cite{wang2023bitnet} as another model compression strategy. The quantization operations are implemented using a PyTorch framework BitNet~\cite{wang2023bitnet}.

\subsubsection*{1-Bit Quantization}
\begin{figure}[h!]
    \centering
    \includegraphics[width=0.45\linewidth]{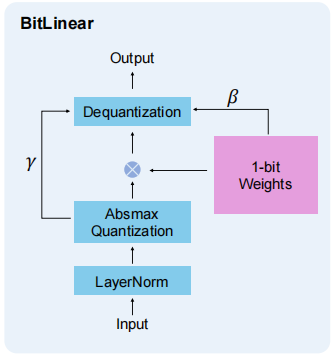}
    \caption{The architecture of BitLinear layer~\cite{wang2023bitnet}.}
    \label{fig:bitlinear}
\end{figure}

In 1-bit quantization, the weight matrix only contains $+1$ or $-1$. Compared to traditional precision like FP16, which requires both multiplication and addition during matrix calculation, 1-bit converts the calculation to addition only. The quantization and dequantization process is presented in Figure~\ref{fig:bitlinear}.

For weight matrix quantization, the weights are first centralized to zero-mean, as shown in Equation~\ref{eq:weight_centering}.
\begin{equation}
\alpha = \frac{1}{nm} \sum_{i,j} W_{ij}, \quad
W' = W - \alpha
\label{eq:weight_centering}
\end{equation}

Then the weights are binarized by applying the sign function. To restore the original scale, a scaling factor $\beta$ is multiplied after quantization (Equation~\ref{eq:scaling_factor}). This dequantizes the binary weights and approximates the real values (Equation~\ref{eq:binarization}):
\begin{equation}
\beta = \frac{1}{nm} \sum_{i,j} \left| W_{ij} \right|
\label{eq:scaling_factor}
\end{equation}
\begin{equation}
\widetilde{W}_{ij} = \beta \cdot \text{sign}(W_{ij}') 
= \beta \cdot \text{sign}(W_{ij} - \alpha)
\label{eq:binarization}
\end{equation}

Absmax quantization~\cite{dettmers2022gpt3} is adopted for activations. The quantization and dequantization process is illustrated in Equation~\ref{eq:absmax}. During quantization, each activation vector is mapped to the range from $-127$ to $127$ by a scaling factor $\gamma = \frac{127}{\|\mathbf{X}\|_\infty}$, where $\|\mathbf{X}\|_\infty = \max_i |X_i|$, and rounded to the nearest integer. Then the dequantized approximation is computed by multiplying the quantized value with $1/\gamma$.
\begin{equation}
\mathbf{X}_{\text{q}} = \text{round} \left( \frac{127}{\max |\mathbf{X}|} \cdot \mathbf{X} \right), \quad
\mathbf{X}_{\text{deq}} = \frac{\max |\mathbf{X}|}{127} \cdot \mathbf{X}_{\text{q}}.
\label{eq:absmax}
\end{equation}

In addition, the straight-through estimator (STE)~\cite{yin2019understanding} is used during training to handle non-differentiable functions like sign and round. In the forward pass, both activations and weights are quantized and then dequantized to simulate the low-precision calculation. For backward pass, STE approximates the quantization operation as an identity function. This allows gradients to be computed only with the real values and bypasses the non-differentiable functions, thus enabling feasible training.

\subsubsection*{Training Workflow}
\begin{figure}[h!]
    \centering
    \includegraphics[width=0.7\linewidth]{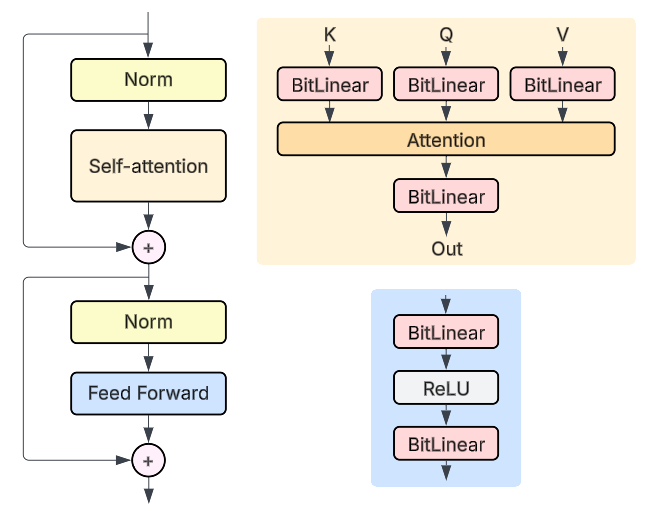}
    \caption{Illustration of replacing standard Linear layers with BitLinear layers in the JetFormer architecture.}
    \label{fig:bitmodel}
\end{figure}
In this training workflow, quantization-aware training (QAT) is used to train the model from scratch. Several linear layers in the JetFormer are replaced with BitLinear layers. As shown in Figure~\ref{fig:bitmodel}, these linear layers include the linear projections in the self-attention blocks (e.g., query, key, value and output layers), and the feedforward layers in the transformer encoder. Other linear layers such as the input embedding and the dense output layer in the classification head are preserved to produce an accurate probability output and maintain the model accuracy. Each weight in the BitLinear layer is quantized to 1-bit during training, which means the weight is either $+1$ or $-1$. To preserve training stability and prevent a large accuracy drop, the activations in the linear layer are kept at 8-bit precision. Unlike the post-training quantization, where the quantization is applied to a trained model, the 1-bit quantized model is trained directly from scratch. This allows the model to adapt to the low precision during training and therefore better maintain the model performance.

At a lower precision, training the model becomes more challenging, especially for the attention layers. The OneCycleLR scheduler used in the previous training experiments is no longer suitable, since frequent changes in the learning rate cause oscillations in the training loss and hinder convergence. Therefore, the ReduceLROnPlateau scheduler is employed in this case, which reduces the learning rate only when the validation loss stops decreasing. The scheduler is set with $\text{factor}=0.8$, $\text{patience}=5$ and $\text{min\_lr}=10^{-4}$. The quantized model is trained for 80 epochs with an initial learning rate of $8\times10^{-4}$. The same JetFormer structure described in Section~\ref{sec:training_settings} is used for the 8-, 16- and 32-particle datasets to compare with the original model performance.

\section{FPGA Implementation}\label{sec:allo}
Having optimized and compressed JetFormer, the next challenge is to deploy it on FPGA hardware. This section presents the hardware workflow using the Allo framework~\cite{chen2024allo}. We begin with an introduction to Allo and its compilation workflow, and then explain the implementation. A particle MLP model is first converted to hardware to validate feasibility. The section then describes the extensions made to Allo to support additional transformer operations, followed by the translation of JetFormer-tiny. Finally, evaluation metrics and potential hardware optimizations are discussed.

\subsection{Optimization}
When developing the models using Allo, several issues occur due to the lack of support for some operations used in the JetFormer-tiny model. To verify the feasibility of the deployment, the particle MLP model is tested first. Unsupported operations required by the JetFormer model are then added, with some modifications to the model implementation. The models are validated using a staged validation pipeline. First, the LLVM CPU backend is used to quickly verify functional correctness. Then, \texttt{sw\_emu} is run to check that the HLS C code produces the correct computational results. Finally, \texttt{hw\_emu} is used to simulate the hardware behavior of the design on an FPGA emulator. Both the original and pruned JetFormer-tiny models are developed. However, since Allo currently lacks support for quantized model deployment on the Vitis HLS backend, models with 1-bit quantization are not deployed in this work.

\subsection{Allo Extension}
\textbf{Frontend limitations.} During the conversion of the JetFormer model to Allo, several issues arise due to missing support for common transformer operations in the existing PyTorch frontend. In particular, operations related to the class token, such as dimension expansion, concatenation of the input embeddings and class token extraction are not supported. Additionally, some operators including log softmax, ReLU and batch normalization with 3D inputs are only partially implemented or unsupported across required dimensions. These limitations result in compilation failures when translating the model to hardware.

\textbf{Support.} To enable full model conversion, both the frontend parsing logic and the kernel-level operator support are extended in Allo. This includes implementing the corresponding compute logic and scheduling strategies for the missing operations in the DSL. Key additions include support for \texttt{log\_softmax}, \texttt{relu3d}, \texttt{batchnorm1d} (for 2D and 3D input), and class token operations such as expansion, concatenation, and slicing. These extensions ensure compatibility with Allo’s compilation flow and allow the JetFormer model to successfully fit on the target hardware platforms.

\subsection{Particle MLP}
\label{sec:mlp_deploy}

The particle MLP model is implemented based on the reference model provided in the \texttt{hls4ml}~\cite{fastml_hls4ml} documentation. It consists of three hidden layers with 64, 32, and 32 neurons followed by an output layer with 5 neurons. All the activation functions are ReLU. The model is trained on the 8-particle 3-feature dataset. The training process follows the same procedure as the JetFormer model described in Section~\ref{sec:training}. Training is conducted with a batch size of 128 for 25 epochs. The optimizer and learning rate scheduler are Adam and CosineAnnealingLR, respectively. The trained model is directly loaded into Allo, with a fixed batch size of 32.

\subsection{JetFormer}
The original and the 50\% compressed JetFormer models are both loaded into Allo for conversion. The JetFormer-tiny model selected from Section~\ref{sec:hpo} has 4 transformer blocks, with an embedding dimension of 8 and 2 attention heads. The model requires 26,168 FLOPs and has 3,080 model parameters, while the compressed version reduces this to 13,784 FLOPs and 1,997 parameters (detailed results are presented in Section~\ref{sec:pruning_results}. In order to fit the full model without exceeding the FPGA computation constraints, a small batch size of 16 is used. Both model accuracy and hardware performance are evaluated after each conversion, and the results are reported in Section~\ref{sec:jetformer_deploy_result}.

\section{Evaluation}
\label{sec:results}
This section illustrates and analyzes the experimental results of the evaluations described in Section~\ref{sec:jetformer} and Section~\ref{sec:allo}. It provides a comprehensive evaluation of JetFormer and the proposed pipeline. The aim is to demonstrate both model performance and the effectiveness of the optimization, compression and deployment strategies. We first analyze JetFormer’s performance on multiple datasets and the outcomes of the hyperparameter optimization study. The impact of structured pruning and 1-bit quantization is then assessed. Finally, the deployment experiments are evaluated using hardware metrics including hardware resource utilization and latency. 

\subsection{Dataset}
Two datasets used for model training and evaluation are described as follows.

\textbf{HLS4ML LHC Jet dataset (150 particles)}~\cite{pierini_2020_3602260}: This dataset contains 620,000 training and 260,000 test samples with balanced classes (q, g, W, Z, t). Each jet has up to 150 particles, each described by 16 kinematic features capturing absolute quantities (momentum components, energy, angles) and relative quantities (relative energy, angular separations, distances to the parent jet)~\cite{sun2025fast}. Feature normalization is performed using Welford’s online algorithm~\cite{welford1962note} to compute mean and standard deviation on the training set only, providing numerically stable, memory-efficient scaling that is then applied to validation and test sets to avoid data leakage.
    
\textbf{\textsc{JetClass} dataset}~\cite{qu2022particle}: \textsc{JetClass} comprises 100M jets (100M/5M/20M for train/validation/test), about two orders of magnitude larger than previous benchmarks, with 10 jet categories and 17 per-particle features including kinematics, PID information, and trajectory displacements, plus additional interaction features. To keep the architecture deployable, JetFormer uses all 17 standard features but excludes the pairwise interaction features. Because of its scale and complexity, only a subset of \textsc{JetClass} is used for training and evaluation. Normalization is the same as ParT using \textsc{JetClass} dataset’s provided statistics (mean and variance), and long-tailed features are log-transformed for numerical stability; zero-padding is applied as in the 150-particle dataset.

\subsection{JetFormer Evaluation}
\subsubsection{Model Performance}
\label{sec:training_settings}
\subsubsection*{Results on the HLS4ML 150-Particle Dataset}
The JetFormer model trained on the HLS4ML 150P dataset uses a compact architecture with 3 transformer encoder layers, an embedding dimension of 64, and 2 attention heads. Notably, we found that dropout could be safely set to 0.0 without harming generalization, likely due to the relatively small input size and strong inductive biases of the architecture. This configuration achieves higher accuracy than previous models across multiple jet tagging benchmarks.

The model is compared with three other models including multilayer perceptron (MLP), deep sets (DS) and interaction network (IN) from Ref.~\cite{odagiu2024ultrafast}. To make a fair comparison, all methods are evaluated on idential 8, 16, and 32 particle datasets with 3 input features, constructed by truncating events from the original HLS4ML dataset (150 particles, 16 features). The results are shown in Table~\ref{tab:result3f}. To benchmark against the baseline model, JEDI-net~\cite{moreno2020jedi}, we also use datasets with 30, 50, 100 and 150 particles and the full set of 16 features (Table~\ref{tab:result16f}).

\begin{table}
\centering
\caption{Summary of model parameters, FLOPs, accuracy and AUC for gluon, light quark, W boson, Z boson and top quark. Model trained on 3-feature dataset with different number of constituents. The results of MLP/DS/IN are from~\cite{odagiu2024ultrafast}.}
\footnotesize
\begin{tabularx}{\textwidth}{@{}ccrrrrrrrrr@{}}
\toprule
\multirow{2}{*}{\textbf{Constituents}} & 
\multirow{2}{*}{\textbf{Model}} & 
\multirow{2}{*}{\textbf{Params}} & 
\multirow{2}{*}{\textbf{FLOPs}} & 
\multirow{2}{*}{\textbf{Accuracy}} & 
\multicolumn{5}{c}{\textbf{AUC}} \\
\cmidrule(lr){6-10}
 & & & & & \textbf{Gluon} & \textbf{Light} & \textbf{W} & \textbf{Z} & \textbf{Top} \\
\midrule
\multirow{4}{*}{8} 
& MLP~\cite{odagiu2024ultrafast}             & 26,826   & 53k   & 64.6 & 0.84 & 0.88 & 0.90 & 0.88 & 0.92 \\
& DS~\cite{odagiu2024ultrafast}              & 3,461    & 37k   & 64.0 & 0.84 & 0.88 & 0.90 & 0.88 & 0.92 \\
& IN~\cite{odagiu2024ultrafast}              & 3,347    & 37k   & 64.9 & 0.84 & 0.88 & 0.91 & 0.89 & 0.92 \\
& JetFormer  & 101,291  & 933k  & \textbf{67.1} & 0.86 & 0.89 & 0.92 & 0.91 & 0.93 \\
\midrule
\multirow{4}{*}{16} 
& MLP~\cite{odagiu2024ultrafast}             & 20,245   & 40k   & 68.4 & 0.87 & 0.89 & 0.91 & 0.90 & 0.94 \\
& DS~\cite{odagiu2024ultrafast}              & 3,461    & 71k   & 69.4 & 0.87 & 0.89 & 0.93 & 0.92 & 0.94 \\
& IN~\cite{odagiu2024ultrafast}              & 3,347    & 140k  & 70.8 & 0.88 & 0.90 & 0.94 & 0.92 & 0.94 \\
& JetFormer  & 102,539  & 2M    & \textbf{74.4} & 0.91 & 0.91 & 0.95 & 0.94 & 0.95 \\
\midrule
\multirow{4}{*}{32} 
& MLP~\cite{odagiu2024ultrafast}             & 24,101   & 48k   & 66.2 & 0.90 & 0.89 & 0.89 & 0.88 & 0.94 \\
& DS~\cite{odagiu2024ultrafast}              & 3,461    & 140k  & 75.9 & 0.91 & 0.91 & 0.96 & 0.95 & 0.95 \\
& IN~\cite{odagiu2024ultrafast}              & 7,400    & 110k  & 75.8 & 0.91 & 0.91 & 0.96 & 0.95 & 0.95 \\
& JetFormer  & 107,339  & 4M    & \textbf{79.9} & 0.94 & 0.93 & 0.97 & 0.97 & 0.97 \\
\bottomrule
\end{tabularx}
\label{tab:result3f}
\end{table}

\begin{table}
\setlength{\tabcolsep}{4.5pt}
\centering
\caption{Summary of model parameters, FLOPs, accuracy and AUC. Model trained on 16-feature dataset with different number of constituents. The results of JEDI-net are from~\cite{moreno2020jedi}.}
\footnotesize
\begin{tabularx}{\textwidth}{@{}lcrrcrccccc@{}}
\toprule
\multirow{2}{*}{\textbf{Model}} & 
\multirow{2}{*}{\textbf{Consti.}} & 
\multirow{2}{*}{\textbf{Params}} & 
\multirow{2}{*}{\textbf{FLOPs}} & 
\multirow{2}{*}{\textbf{Accuracy}} & 
\multicolumn{5}{c}{\textbf{AUC}} \\
\cmidrule(lr){6-10}
& & & & & \textbf{Gluon} & \textbf{Light} & \textbf{W} & \textbf{Z} & \textbf{Top} \\
\midrule
JEDI-net~\cite{moreno2020jedi}                  & 100 & 33,625   & 116M  & \textbackslash{} & 95.29 & 93.01 & 97.39 & 96.79 & 96.83 \\
\midrule
JEDI-net w/ $\sum\mathcal{O}$~\cite{moreno2020jedi} & 150 & 8,767    & 458M  & \textbackslash{} & 95.28 & 92.90 & 96.95 & 96.49 & 96.77 \\
\midrule
\multirow{4}{*}{JetFormer} 
                          & 30  & 107,403  & 4M    & 81.77 & 95.34 & 93.44 & 97.68 & 97.36 & 97.15 \\
                          & 50  & 117,243  & 6M    & 82.88 & 95.90 & 93.88 & 98.09 & 97.80 & 97.43 \\
                          & 100 & 162,843  & 14M   & 83.05 & 95.99 & 93.92 & 98.12 & 97.83 & 97.44 \\
                          & 150 & 238,443  & 24M   & 82.97 & 95.96 & 93.89 & 98.10 & 97.81 & 97.42 \\
\bottomrule
\end{tabularx}
\label{tab:result16f}
\end{table}

The results show that the JetFormer model achieves consistently better performance across all datasets. The accuracy improves by approximately 3\% to 4\%, with higher AUC scores for each jet class. However, the size of JetFormer is significantly larger compared to other models, especially on a larger dataset. Therefore, model compression is necessary for later hardware deployment in order to reduce computational cost and inference latency.

\subsubsection*{Results on \textsc{JetClass} Dataset}

For the \textsc{JetClass} dataset, we scale up JetFormer to a larger configuration with 10 transformer encoder layers, an embedding dimension of 128, and 8 attention heads. As with the smaller variant, we set dropout to 0.0, which we found sufficient for stable training and strong generalization on this task. 

The model is trained and validated on the subset with 2M samples each, and the results are reported in Table~\ref{tab:jetclass_result}. Without the additional pairwise interaction features considered in ParT~\cite{qu2022particle} model, JetFormer can achieve a competitive result, with only 0.7\% and 0.07\% decreases in accuracy and AUC. At the same time, JetFormer requires 37.4\% fewer FLOPs compared to ParT, which indicates its computational efficiency. Training on the 2M-sample subset takes approximately three days on a single NVIDIA GeForce RTX 3090 GPU. Furthermore, since JetFormer can generalize effectively across both small benchmark datasets and the large-scale \textsc{JetClass} dataset, it demonstrates strong scalability and is practical for downstream hardware deployment.

\begin{table}
\centering
\caption{Model performance on \textsc{JetClass} dataset. All models are trained on the 2M subset. The results of ParticleNet and ParT are from~\cite{qu2022particle}.}
\begin{tabular}{lcccc}
\toprule
 & \textbf{Accuracy} & \textbf{AUC} & \textbf{Params} & \textbf{FLOPs} \\
\midrule
ParticleNet & 0.828 & 0.9820 & 370k & 540M \\
ParT        & 0.836 & 0.9834 & 2.14M & 340M \\
JetFormer   & 0.829 & 0.9827 & 1.66M & 213M \\
\bottomrule
\end{tabular}
\label{tab:jetclass_result}
\end{table}

\subsubsection{Hyperparameter Optimization}
\subsubsection*{Sampler Selection}
\begin{figure}[h!]
    \centering
    \includegraphics[width=1\linewidth]{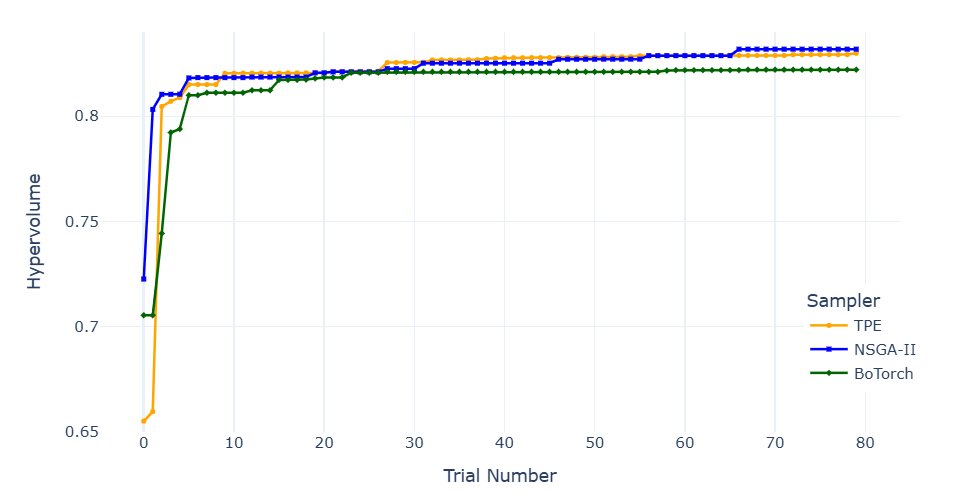}
    \caption{Hypervolume of TPESampler, NSGAIISampler, and BoTorchSampler over 80 valid trials.}
    \label{fig:samples}
\end{figure}

The hypervolume curve shown in Figure~\ref{fig:samples} compares the performance of TPESampler, NSGAIISampler and BoTorchSampler over 80 valid trials. From the plot, all three samplers finally reach a comparable hypervolume of around 0.83. NSGAII achieves a high hypervolume early and continues to improve steadily, which achieves the highest result. It shows a faster convergence and consistent exploration of the hyperparameter space. TPE starts slower but eventually approximates NSGAII’s performance. BoTorch lags slightly behind both in the early and later phases, which may be less effective for this hyperparameter search space. Therefore, NSGAIISampler is more reliable and is selected as the final optimization sampler.

\subsubsection*{Best Trials}
\begin{figure}[h!]
    \centering
    \includegraphics[width=0.7\linewidth]{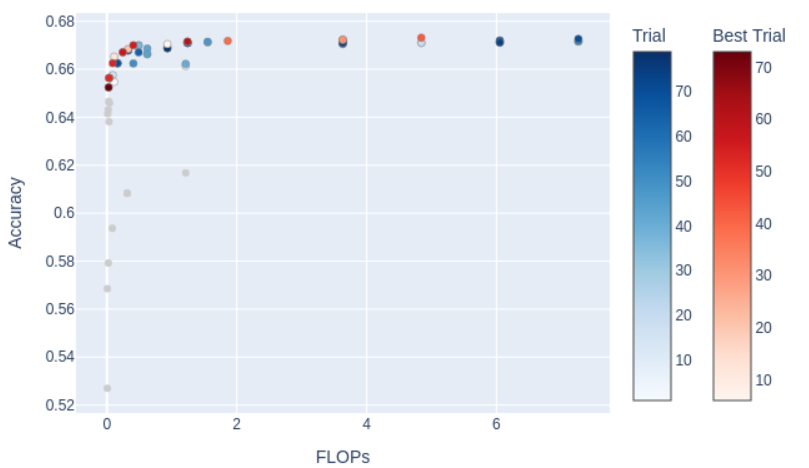}    \caption{The Pareto front plot using NSGAIISampler on 8-particle 3-feature dataset.}
    \label{fig:pareto}
\end{figure}
The Pareto front plot of the HPO process with the NSGAIISampler on the dataset with 8 particles and 3 features is shown in Figure~\ref{fig:pareto}. Red points indicate optimal trials, while gray points represent infeasible solutions with accuracy below 0.65. Table~\ref{tab:trials} shows all the best trials identified among 80 runs, ranked by accuracy from lowest to highest. The validation accuracy generally increases with the number of FLOPs. Since the first trial (Model 0) has the fewest FLOPs while still achieving a vlidation accuracy above 0.65, it is chosen as JetFormer-tiny for subsequent model compression and deployment process. The final test accuracy for this model achieves 0.656. With a comprehensive search, the best model, consisting of 4 transformer blocks with an embedding dimension of 128, 8 attention heads and a dropout rate of 0.05, can achieve a test accuracy of 0.673.

\begin{table}
\centering
\caption{Best trials among 80 runs using NSGAIISampler.}
\begin{tabular}{c ccccccc}
\toprule
\textbf{Index} & \textbf{FLOPs} & \textbf{val\_acc} & \textbf{\# transformers} & \textbf{embed\_dim} & \textbf{\# heads} & \textbf{dropout} \\
\midrule
0 & 26,168 & 0.6525 & 4 & 8   & 2 & 0.00 \\
1 & 32,648 & 0.6564 & 5 & 8   & 2 & 0.00 \\
2 & 39,128 & 0.6564 & 6 & 8   & 2 & 0.00 \\
3 & 89,200 & 0.6625 & 4 & 16  & 2 & 0.00 \\
4 & 111,376 & 0.6653 & 5 & 16  & 2 & 0.00 \\
5 & 244,640 & 0.6671 & 3 & 32  & 2 & 0.00 \\
6 & 325,856 & 0.6685 & 4 & 32  & 2 & 0.00 \\
7 & 407,072 & 0.6700 & 5 & 32  & 2 & 0.00 \\
8 & 931,654 & 0.6705 & 3 & 64  & 4 & 0.00 \\
9 & 1,241,544 & 0.6715 & 4 & 64  & 2 & 0.00 \\
10 & 1,861,324 & 0.6718 & 6 & 64  & 4 & 0.05 \\
11 & 3,632,780 & 0.6724 & 3 & 128 & 8 & 0.00 \\
12 & 4,842,384 & 0.6732 & 4 & 128 & 8 & 0.05 \\
\bottomrule
\end{tabular}
\label{tab:trials}
\end{table}

\subsubsection{Structured Pruning}
\label{sec:pruning_results}
The 50\% pruning result is shown in Table~\ref{tab:pruning_results}. For the JetFormer-tiny (Model 0), FLOPs and the number of parameters are reduced to 13.78k and 1,997 respectively, with only around 0.5\% loss in accuracy. For this smallest model, the GPU inference time with batch size of 10,240 decreases by 17.46\%. It is easier for larger models to maintain the accuracy after pruning. The accuracy loss for these models is less than 0.3\%. Due to space limitation, only the performance of Model 1 and Model 2 is listed here.

\begin{table}
\centering
\caption{Model performance before and after structured pruning using best models from Table~\ref{tab:trials}.}
\begin{tabular}{l l c c c}
\toprule
\textbf{Index} & \textbf{Metric} & \textbf{Before} & \textbf{After} & \textbf{Change (\%)} \\
\midrule
\multirow{4}{*}{\textbf{Model 0}} 
 & FLOPs         & 26,168 & 13,784 & 47.32 \\
 & Params        & 3,085  & 1,997  & 35.27 \\
 & Accuracy      & 0.656  & 0.653  & 0.49 \\
 & Inference time & 3.517 $\pm$ 0.114 ms & 2.902 $\pm$ 0.008 ms & 17.46 \\
\midrule
\multirow{4}{*}{\textbf{Model 1}} 
 & FLOPs         & 32,648 & 17,168 & 47.41 \\
 & Params        & 3,831  & 2,471  & 35.50 \\
 & Accuracy      & 0.656  & 0.655  & 0.28 \\
 & Inference time & 4.331 $\pm$ 0.123 ms & 3.575 $\pm$ 0.007 ms & 17.46 \\
\midrule
\multirow{4}{*}{\textbf{Model 2}} 
 & FLOPs         & 39,128 & 20,552 & 47.47 \\
 & Params        & 4,577  & 2,945  & 35.66 \\
 & Accuracy      & 0.659  & 0.658  & 0.17 \\
 & Inference time & 4.904 $\pm$ 0.248 ms & 4.060 $\pm$ 0.006 ms & 17.21 \\
\bottomrule
\end{tabular}
\label{tab:pruning_results}
\end{table}

\subsubsection{1-Bit Quantization}
The quantized model uses the same JetFormer structure with 3 transformer blocks, 2 attention heads, an embedding dimension of 64 and no dropout. Table~\ref{tab:quantization_results} summarizes the comparison between the original and the 1-bit quantized models. The results show that the quantized model achieves a significant reduction in model size of over 82–92\%, while the accuracy drop remains relatively small at only 1.5–3.5\%. As the number of particles increases, the relative accuracy drop becomes larger, reaching 3.5\% for 32 constituents. This means the quantization error may accumulate more significantly for higher-dimensional inputs. Although there is a slight performance drop, the overall compression is substantial. The quantized model is much more practical for deployment and real-world applications where hardware efficiency is critical. To further improve the quantized model capacity on larger datasets, a more expressive quantization strategy may be required.

\begin{table}
\centering
\caption{Model size and accuracy before and after 1-bit quantization, with relative percentage change.}
\begin{tabular}{ccccc}
\toprule
\textbf{Constituents} & \textbf{Metric} & \textbf{Before} & \textbf{After} & \textbf{Change (\%)} \\
\midrule
\multirow{2}{*}{8}  & Model size (KB) & 404.12 & 31.37 & 92.24 \\
                    & Accuracy        & 0.671  & 0.661 & 1.49  \\
\midrule
\multirow{2}{*}{16} & Model size (KB) & 413.87 & 41.12 & 90.07 \\
                    & Accuracy        & 0.744  & 0.728 & 2.15  \\
\midrule
\multirow{2}{*}{32} & Model size (KB) & 451.37 & 78.62 & 82.58 \\
                    & Accuracy        & 0.799  & 0.771 & 3.50  \\
\bottomrule
\end{tabular}
\label{tab:quantization_results}
\end{table}

\subsection{Hardware Evaluation}
\subsubsection{Hardware Evaluation of Particle MLPs}
The resource utilization of the particle MLP model described in Section~\ref{sec:mlp_deploy} is presented in Table~\ref{tab:mlp_util}. Since Allo does not provide the full pipelining, the latency of this model is 0.585 ms at a batch size of 32, which is relatively high. The utilization of each hardware resource is less than 1\%, indicating significant potential for latency reduction through improved resource utilization.
\begin{table}
\centering
\caption{FPGA resource utilization of MLP particle model with batch size of 32.}
\begin{tabular}{lcccc}
\toprule
 & \textbf{BRAM 18K} & \textbf{DSP} & \textbf{FF} & \textbf{LUT} \\
\midrule
Total used    & 30     & 5     & 8,191     & 11,792 \\
Available     & 5,376  & 12,288 & 3,456,000 & 1,728,000 \\
Utilization (\%) & 0.56 & 0.04  & 0.24      & 0.68 \\
\bottomrule
\end{tabular}
\label{tab:mlp_util}
\end{table}

\subsubsection{Hardware Evaluation of JetFormer}
\label{sec:jetformer_deploy_result}
The resource utilization of JetFormer-tiny model before and after compression is shown in Table~\ref{tab:jetformer_util}. The latency of the original and pruned models are 4.767 ms and 2.705 ms with a batch size of 16. The results show the pruned model achieves lower latency and reduced resource utilization, although the overall latency is still high without pipelining. With a batch size of 2, the latency of the pruned model can be reduced to 0.404 ms, with a further reduction in resource utilization.

Please note that no dedicated hardware optimizations were applied in this implementation. The design was generated using Allo for HLS code generation, without manual pipelining, loop unrolling, or other architecture-specific enhancements. As evidenced by the low FPGA resource utilization reported in Table~\ref{tab:jetformer_util}, which remains well below the available capacity, there is substantial room for performance improvement. In particular, increasing parallelism, for example through deeper pipelining or spatial replication, could significantly reduce latency while making more efficient use of the abundant on-chip resources. We leave this for our future work.

\begin{table}
\centering
\caption{FPGA resource utilization before and after pruning for the JetFormer-tiny model with batch size of 16 and 2 (bottom).}
\begin{tabular}{l l c c c c}
\toprule
 &  & \textbf{BRAM 18K} & \textbf{DSP} & \textbf{FF} & \textbf{LUT} \\
\midrule
 & Available & 5,376 & 12,288 & 3,456,000 & 1,728,000 \\
\midrule
\multirow{2}{*}{Original} 
 & Total used    & 511 & 131 & 107,847 & 139,143 \\
 & Utilization (\%) & 9.51 & 1.07 & 3.12 & 8.05 \\
\midrule
\multirow{2}{*}{Pruned} 
 & Total used    & 527 & 91  & 95,527  & 129,554 \\
 & Utilization (\%) & 9.80 & 0.74 & 2.76 & 7.50 \\
\midrule
\multirow{2}{*}{\makecell[l]{Pruned \\ (batch size 2)}} 
 & Total used    & 157 & 75  & 88,513  & 120,041 \\
 & Utilization (\%) & 2.92 & 0.61 & 2.56 & 6.95 \\
\bottomrule
\end{tabular}
\label{tab:jetformer_util}
\end{table}

\subsubsection{Discussion}
\label{sec:potential_mlp}
Table~\ref{tab:with_pipelining} illustrates the results of the two-layer MLP model with and without applying full pipelining. After full pipelining, the resource utilization increases significantly, while the latency drops from 7.439 us to 1.662 us. This indicates an improvement direction in transformer model acceleration. Additional pipelining methods may be required to speed up transformer models while keeping the resource utilization within the hardware constraints.
\begin{table}
\centering
\caption{Resource utilization comparison (with and without full pipelining) for the 2-layer MLP model.}
\begin{tabular}{l l c c c c}
\toprule
 &  & \textbf{BRAM 18K} & \textbf{DSP} & \textbf{FF} & \textbf{LUT} \\
\midrule
 & Available & 4,032 & 9,024 & 2,607,360 & 1,303,680 \\
\midrule
\multirow{2}{*}{Default}
 & Total used       & 2 & 21   & 7,781   & 10,112 \\
 & Utilization (\%) & 0.05 & 0.23 & 0.30 & 0.78 \\
\midrule
\multirow{2}{*}{Full pipeline}
 & Total used       & 0 & 6,877 & 757,888 & 399,696 \\
 & Utilization (\%) & 0.00 & 76.21 & 29.07 & 30.66 \\
\bottomrule
\end{tabular}
\label{tab:with_pipelining}
\end{table}

\section{Conclusion and Future Work}
\label{sec:conclusion}

We presented JetFormer, a scalable encoder-only transformer for jet tagging that serves both high-accuracy offline analysis and low-latency online triggering within a unified architecture. On the \textsc{JetClass} dataset, full-scale JetFormer matches ParT’s accuracy (within 0.7\%) with 37.4\% fewer FLOPs, while compact variants, optimized via multi-objective search, pruning, and 1-bit quantization, enable FPGA deployment under stringent latency constraints.
Future work includes further hardware optimization to reduce the inference latency of FPGA-based JetFormer implementation, advancing co-design strategies that jointly optimize models and hardware architectures, and extending JetFormer to other HEP tasks. 

\FloatBarrier

\noindent \textbf{Acknowledgement:} 

The support from the United Kingdom EPSRC ( EP/V028251/1, EP/N031768/1, EP/S030069/1, and EP/X036006/1), UKRI256 and STFC (grant number ST/W000636/1), Schmidt Sciences. A.G., and U.S. Department of Energy under contract number DE-AC02-76SF00515, 
Intel and Lenovo (ICICLE HPC \& AI partnership) and AMD is gratefully acknowledged. 

\bibliographystyle{ieeetr}
\bibliography{bibliography}

\end{document}